\documentclass[10pt,twocolumn,letterpaper]{article}

\usepackage{iccv}
\usepackage{times}
\usepackage{epsfig}
\usepackage{graphicx}
\usepackage{amsmath}
\usepackage{amssymb}

\usepackage[breaklinks=true,bookmarks=false]{hyperref}

\iccvfinalcopy 

\def\iccvPaperID{14} 

\ificcvfinal\fi

\begin{document}

\title{5th Place Solution for VSPW 2021 Challenge}

\author{Jiafan Zhuang, Yixin Zhang, Xinyu Hu, Junjie Li\\
	University of Science and Technology of China\\
	{\tt\small \{jfzhuang, zhyx12, yuanlaishimeng, hnljj\}@mail.ustc.edu.cn}
	\and
	Zilei Wang\\
	University of Science and Technology of China\\
	{\tt\small zlwang@ustc.edu.cn}
}

\maketitle
\ificcvfinal\thispagestyle{empty}\fi

\begin{abstract}
	In this article, we introduce the solution we used in the VSPW 2021 Challenge. Our experiments are based on two baseline models, Swin Transformer and MaskFormer. To further boost performance, we adopt stochastic weight averaging technique and design hierarchical ensemble strategy. Without using any external semantic segmentation dataset, our solution ranked the 5th place in the private leaderboard. Besides, we have some interesting attempts to tackle long-tail recognition and overfitting issues, which achieves improvement on val subset. Maybe due to distribution difference, these attempts don't work on test subset. We will also introduce these attempts and hope to inspire other researchers.
\end{abstract}

\section{Introduction}

Semantic segmentation aims to assign a unique semantic label to every pixel in a given image, which is a fundamental research topic in the computer vision community and has many potential applications such as image editing, autonomous driving and robotics. To further aid the development of semantic segmentation, ~\cite{miao2021vspw} presents a large-scale video scene parsing dataset, called VSPW dataset. VSPW dataset is split into three subsets, containing train subset with 2806 videos, val subset with 343 videos and test subset with 387 videos. Each video  contains 11 to 241 frames. VSPW dataset totally annotates 3536 videos, including 251,633 frames from 124 categories.

\begin{figure}[t]
	\begin{center}
		\includegraphics[width=1.0\linewidth]{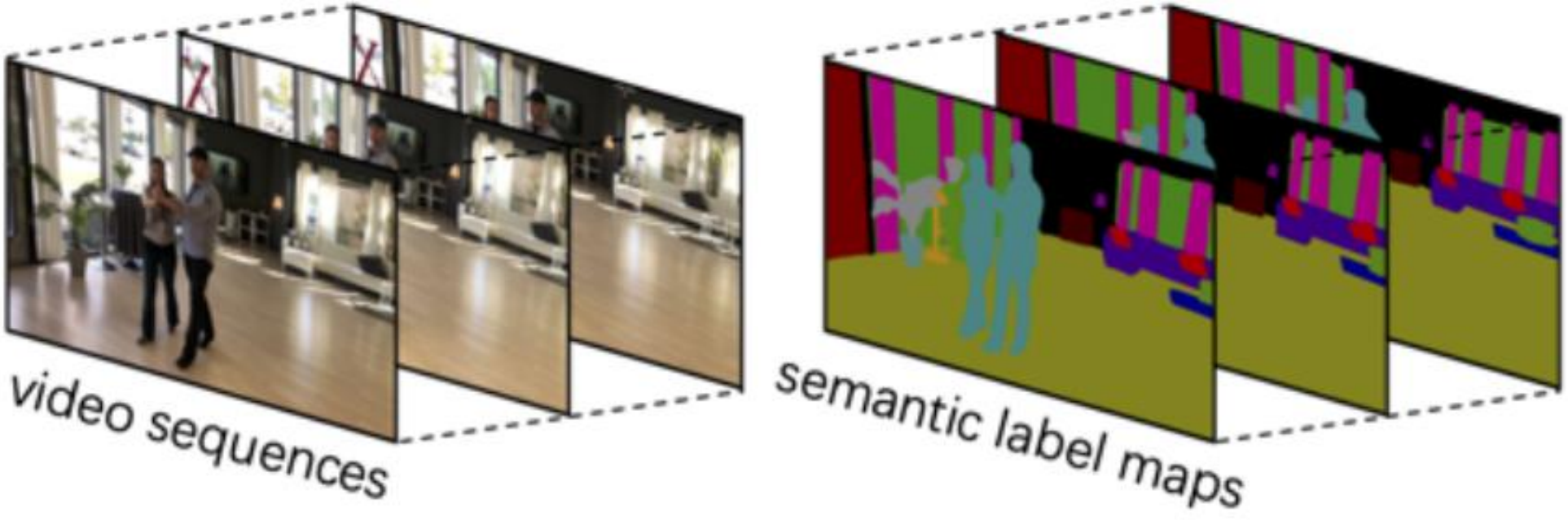}
	\end{center}
	\caption{One example for the video scene annotation.}
	\label{fig:dataset}
\end{figure}

In our dataset analysis, we found that the most distinct characteristic of VSPW dataset is category imbalance. In the train subset, the largest and smallest class has existed in 2110 and only 10 videos, respectively. Therefore, long-tail recognition is a key issue. Besides, since the dataset is composed of videos and consecutive frames in the same video are often similar in a large portion of content, training data has obvious homogenization, which would cause overfitting issue.

To tackle these issues, we have some interesting attempts, including logits adjustment for long-tail recognition and self distillation. To be specific, we integrate the class prior into softmax cross-entropy loss and learning a residual logits during training, which can effectively compensate class distribution difference. Besides, we introduce an extra regularization to penalize the predictive distribution between teacher and student models, which can relieve the overfitting issue. Strangely, our adopted methods can bring improvement on val subset but get even worse performance on test subset, which may indicate that there is distribution difference between val and test subsets.

Besides, to further boost the performance of our baseline models, we adopt stochastic weight averaging and design hierarchical ensemble strategy for better model ensemble performance. Experimental results show that our methods can bring significant improvement. All training is conducted on the VSPW dataset only without any external semantic segmentation dataset. Our solution ranked the 5th place in the private leaderboard. We will elaborate the details of our solution in the following sections.

\section{Method}

\subsection{Baseline Model}

Since transformer-based models achieve excellent performance in semantic segmentation task, we  mainly leverage Swin Transformer~\cite{liu2021swin} and MaskFormer~\cite{cheng2021per} as our baseline models. These two models both achieve top performance on ADE20K~\cite{zhou2017scene} leaderboard. In our experiments, Swin-L model is chosen as the backbone of two baselines, which is pretrained on ImageNet-22k. When tuning the baseline model, we found that large crop size during training phase and OHEM can boost the performance of Swin Transformer, as shown in Table~\ref{tab:baseline}. We set confidence score threshold as 0.7 for hard example selection. Besides, we set the minimum number of predictions to keep as 200k for stable training in the early stage. However, these adjustments have no influence on MaskFormer. To further boost the performance, 
we perform test-time augmentation(TTA) with horizontal filp and multi-scale when submitting to the server for test subset evaluation.

\begin{table}[h]
	\begin{center}
		\begin{tabular}{l|c|c}
			Method				& Val mIoU (\%)	   	& Test mIoU (\%)		\\
			\hline
	        Swin, CS=480  		& 56.13				& 48.63					\\
			Swin, CS=640 		& 56.48				& 49.31					\\
			Swin, CS=640, OHEM 	& 56.94				& 49.73					\\
			MaskFormer			& 56.52				& 53.99					\\
			\hline
		\end{tabular}
	\end{center}
	\caption{Performance comparison between baseline models under different setting. CS represents for crop size.}
	\label{tab:baseline}
\end{table}

\subsection{Stochastic Weight Averaging}

Stochastic Weight Averaging (SWA) is developed in~\cite{izmailov2018averaging} for improving generalization in deep networks. In~\cite{zhang2020swa}, SWA was attempted in object detection task and effectively improving the performance of detector. Without any inference cost and any change to the detector, SWA can consistently bring 1.0 AP improvement over various popular detectors on the challenging COCO benchmark.

In our solution, we also adopt SWA to improve the performance of video semantic segmentation. To be specific, after training the baseline model, we train the model for an extra 20k iterations using cyclical learning rates and then save checkpoints every 2k iterations. After that, we average these saved checkpoints as our final segmentation model. As shown in Table~\ref{tab:swa}, SWA can bring consistent improvement over different baseline models.

\begin{table}[h]
	\begin{center}
		\begin{tabular}{l|c}
			Method				& Test mIoU (\%)		\\
			\hline
	        Swin  				& 51.75					\\
			Swin w/ SWA 		& 52.58					\\
			MaskFormer 			& 53.36					\\
			MaskFormer w/ SWA	& 54.00					\\
			\hline
		\end{tabular}
	\end{center}
	\caption{SWA can bring consistent improvement over different baseline models.}
	\label{tab:swa}
\end{table}

\subsection{Hierarchical Ensemble Strategy}

Model ensemble is a popular technique in competitions. The most commonly used method is to average predictions from different models, which may restrict the performance of strong models. In our solution, we design a hierarchical ensemble strategy for better performance. To be specific, we split models into several groups and conduct prediction average in each group respectively. After that, we search optimal weights to fused predictions from different groups according to the performance on val subset.

We first train 11 MaskFormer models and 11 Swin Transformer models under different settings, such as different crop sizes, data splits and random seeds. Since MaskFormer always has better performance than Swin Transformer, we regard MaskFormer models as basic model and select 6 strong Swin Transformer models as auxiliary models $M_{A}$. Further, we split MaskFormer models into two groups, 3 models with higher scores on val subset as strong models $M_{S}$ and 8 weak models $M_{W}$. The prediction of each group is the averaged results of models in that group. Then, we grid search two weights $\alpha$ and $\beta$ to fuse predictions from different groups. Finally, the final prediction is

\begin{equation}
P_{Final} = P_{S} + P_{W} * \alpha + P_{A} * \beta
\end{equation}

In our experiments, we set $\alpha$ as 1.4 and $\beta$ as 1.0 after grid search. As shown in Table~\ref{tab:ensemble}, our proposed hierarchical ensemble strategy can bring significant improvements.

\begin{table}[h]
	\begin{center}
		\begin{tabular}{l|c|c}
			Method								& Val			& Test		\\
			\hline
	        Best Single Model  					& 57.26			& 53.99		\\
			$P_{S}$ 							& 58.01			& 55.10		\\
			$P_{S} + P_{W} * 1.4$ 				& 58.50			& 55.16		\\
			$P_{S} + P_{W} * 1.4 + P_{A} * 1.0$	& 61.10			& 55.48		\\
			\hline
		\end{tabular}
	\end{center}
	\caption{Hierarchical ensemble strategy can bring significant improvement.}
	\label{tab:ensemble}
\end{table}

\section{Interesting Attempts}

We have some interesting attempts and observe obvious improvement on val subset. When conducing experiments on test subset, performance don't get improved and even worse, which may indicate that there is obvious distribution difference between val and test subsets. In the following, we will introduce these interesting attempts and hope to inspire other researchers.

\subsection{Logits Adjustment}
\begin{figure}[!ht]
	\centering
	\includegraphics[width=0.98\columnwidth]{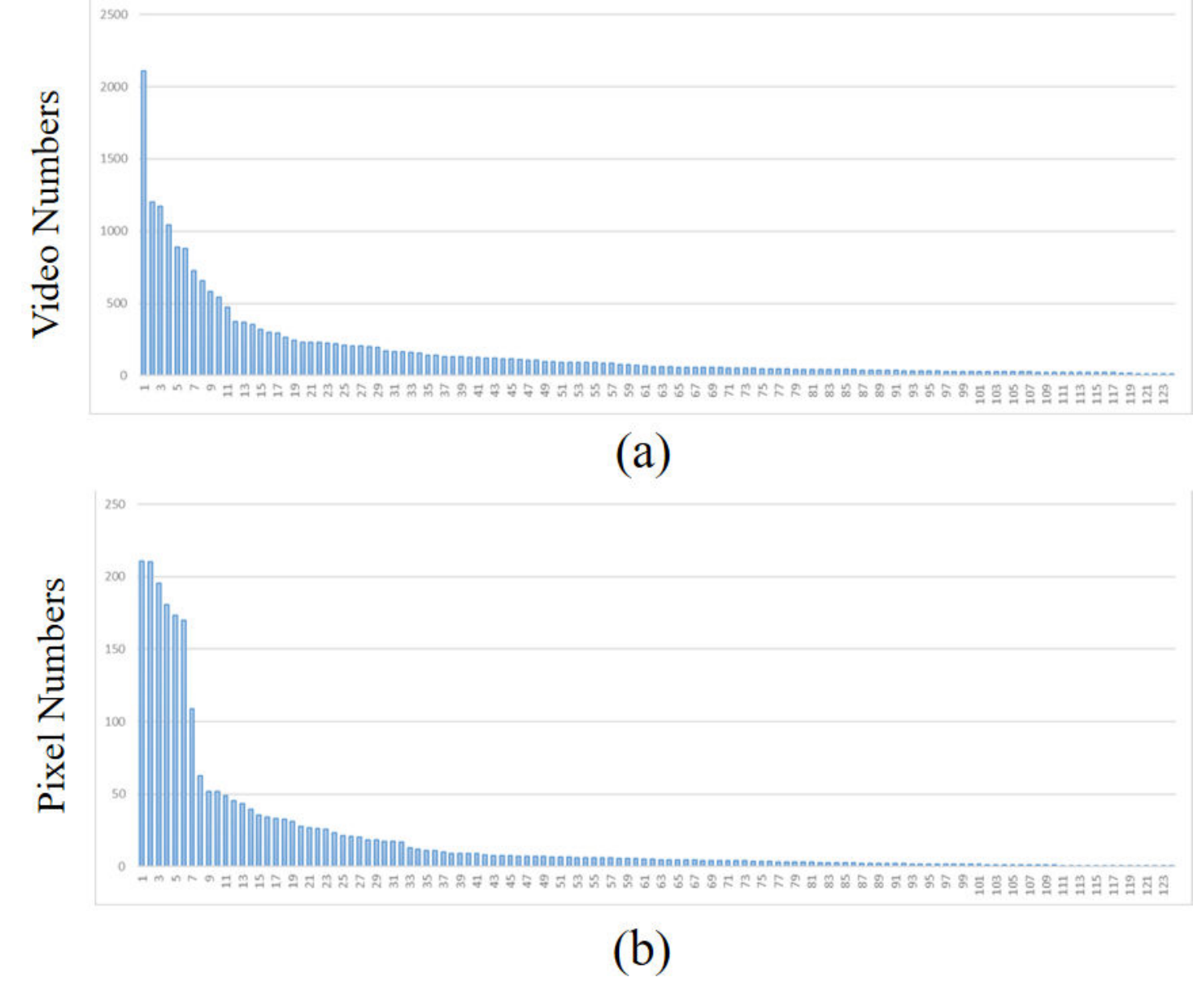}
	\caption{\small (a) Video numbers that each category appears in. (b) Pixel numbers of each category. The results are obtained by entire training dataset. Pixel number is scaled down by $10^{8}$.}
	\label{fig:class_imbalance}
\end{figure}

From Fig~\ref{fig:class_imbalance}, it can be seen that there exists huge imbalance among 
different classes, thus the rare class can not be well learned. Following~\cite{menon2020long}, we address the class-imbalance problem via logits adjusted softmax cross-entropy. The main idea is integrating the class prior into softmax cross-entropy loss and learning a residual logits during training. And during testing, only the learned residual logits is used for prediction. To do so, we consider weight and bias in the final convolutional layer as the learnable part and prior part. 

\begin{equation}
\label{eqn:logit-cal}
\resizebox{0.9\linewidth}{!}{
	$\displaystyle
	logits_{y} =\frac{w_{y} * f(x)}{||w_{y}||_{2} * ||f(x)||_{2}}  + \tau \cdot \log \pi_{y}
	$}
\end{equation}

where $w_{y}$ represents the weight parameter of class $y$ in the final convolutional layer, and $f(x)$ is the extracted feature. Here we adopt the normalized weights and feature to further alleviate the class imbalance problem. $\tau \cdot \log \pi_{y}$ is the bias of class $y$ which is fixed during training. $\pi_{y}$ is the prior probability of class $y$ corresponding to Fig~\ref{fig:class_imbalance}(b). $\tau$ is a scale parameter where we set to 0.03 by hyper-parameter searching in val dataset. Thus, the cross-entropy loss of sample $x$ with label $y$ is calculated as follows:

\begin{equation}
\label{eqn:logit-adjusted-loss}
\resizebox{0.7\linewidth}{!}{
	$\displaystyle
	{\ell}( y, x ) = -\log \frac{e^{logits_{y}}}{\sum_{y' \in [L]} e^{logits_{y'}}}
	$}
\end{equation}

Table~\ref{tab:logits_adjust} shows the experiment results of logits adjusted models. It can be seen that the performance in test subset is not consistent with val subset. We conjecture that the difference of class ratio in val subset and test subset can not be ignored.

\begin{table}[h]
	\begin{center}
		\begin{tabular}{l|c|c}
			Method				& Val mIoU (\%)	   	& Test mIoU (\%)		\\
			\hline
			Swin				& 56.94				& 49.73					\\
			Swin w/ LA			& 57.62				& 49.97					\\
			\hline
			MaskFormer			& 56.52				& 53.99					\\
			MaskFormer w/ LA		& 57.63				& 52.79					\\
			\hline
		\end{tabular}
	\end{center}
	\caption{Performance comparison between baseline models and logits adjusted (LA) models.}
	\label{tab:logits_adjust}
\end{table}

\subsection{Self Distillation}

Deep neural network with millions of parameters may suffer from poor generalization due to overfitting. In our experiments, we observe that accuracy get improved consistently on training subset, but accuracy on val subset get early saturated, which may caused by the overfitting. To mitigate the issue, we adopt a new technique called self distillation~\cite{tian2020rethinking} as an extra regularization. Self distillation is a new regularization method that penalizes the predictive distribution between teacher model and student model.

\begin{figure}[t]
	\begin{center}
		\includegraphics[width=1.0\linewidth]{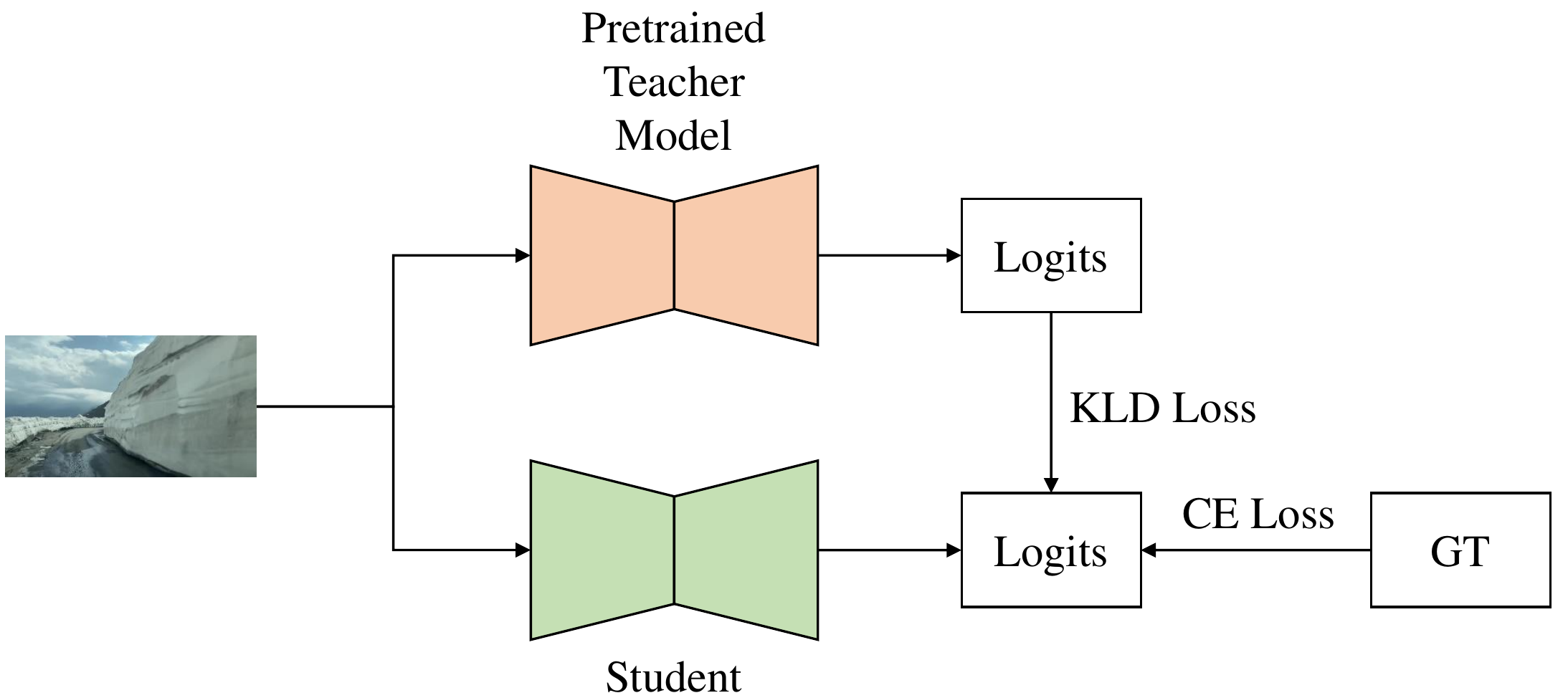}
	\end{center}
	\caption{The illustration of self distillation method.}
	\label{fig:self-distill}
\end{figure}

To be specific, we first train a model and save it as a teacher model. After that, we initialize a student model, which has the same architecture as teacher model. In the training phase, besides original cross entropy loss, we implement a KL distribution loss between logits predicted from teacher and student model as an extra regularization. The framework is shown in Figure~\ref{fig:self-distill}.

The experimental results are shown in Table~\ref{tab:sd}. Obviously, self distillation can bring improvement on val subset but get worse performance on test subset.

\begin{table}[h]
	\begin{center}
		\begin{tabular}{l|c|c}
			Method				& Val mIoU (\%)		& Test mIoU (\%)	\\
			\hline
	        Swin  				& 55.40				& 50.99				\\
			Swin w/ SD 			& 56.26				& 50.52				\\
			\hline
		\end{tabular}
	\end{center}
	\caption{Self Distillation can bring improvement on val subset but get worse performance on test subset.}
	\label{tab:sd}
\end{table}

{\small
	\bibliographystyle{ieee_fullname}
	\bibliography{egbib}

\begin{thebibliography}{1}\itemsep=-1pt

\bibitem{cheng2021per}
Bowen Cheng, Alexander~G Schwing, and Alexander Kirillov.
\newblock Per-pixel classification is not all you need for semantic
  segmentation.
\newblock {\em arXiv preprint arXiv:2107.06278}, 2021.

\bibitem{izmailov2018averaging}
Pavel Izmailov, Dmitrii Podoprikhin, Timur Garipov, Dmitry Vetrov, and
  Andrew~Gordon Wilson.
\newblock Averaging weights leads to wider optima and better generalization.
\newblock In {\em UAI}, 2018.

\bibitem{liu2021swin}
Ze Liu, Yutong Lin, Yue Cao, Han Hu, Yixuan Wei, Zheng Zhang, Stephen Lin, and
  Baining Guo.
\newblock Swin transformer: Hierarchical vision transformer using shifted
  windows.
\newblock {\em arXiv preprint arXiv:2103.14030}, 2021.

\bibitem{menon2020long}
Aditya~Krishna Menon, Sadeep Jayasumana, Ankit~Singh Rawat, Himanshu Jain,
  Andreas Veit, and Sanjiv Kumar.
\newblock Long-tail learning via logit adjustment.
\newblock In {\em ICLR}, 2021.

\bibitem{miao2021vspw}
Jiaxu Miao, Yunchao Wei, Yu Wu, Chen Liang, Guangrui Li, and Yi Yang.
\newblock Vspw: A large-scale dataset for video scene parsing in the wild.
\newblock In {\em CVPR}, 2021.

\bibitem{tian2020rethinking}
Yonglong Tian, Yue Wang, Dilip Krishnan, Joshua~B Tenenbaum, and Phillip Isola.
\newblock Rethinking few-shot image classification: a good embedding is all you
  need?
\newblock In {\em ECCV}, 2020.

\bibitem{zhang2020swa}
Haoyang Zhang, Ying Wang, Feras Dayoub, and Niko S{\"u}nderhauf.
\newblock Swa object detection.
\newblock {\em arXiv preprint arXiv:2012.12645}, 2020.

\bibitem{zhou2017scene}
Bolei Zhou, Hang Zhao, Xavier Puig, Sanja Fidler, Adela Barriuso, and Antonio
  Torralba.
\newblock Scene parsing through ade20k dataset.
\newblock In {\em CVPR}, 2017.

\end{thebibliography}
}

\end{document}